\documentclass{bmvc2k}

\usepackage{booktabs}
\usepackage{amssymb}
\usepackage{listings}
\usepackage{comment}
\usepackage{diagbox}
\usepackage{algorithm}
\usepackage{algpseudocode}

\title{Space-Time Memory Network for Sounding Object Localization in Videos}

\addauthor{Sizhe Li$^{\ast}$}{sli96@u.rochester.edu}{1}
\addauthor{Yapeng Tian$^{\ast}$}{yapengtian@rochester.edu}{1}
\addauthor{Chenliang Xu}{chenliang.xu@rochester.edu}{1}

\addinstitution{
 Department of Computer Science\\
 University of Rochester\\
 Rochester, USA 
}

\runninghead{Li, Tian, Xu}{Sounding Object Localization in Videos}

\begin{document}

\maketitle

\begin{abstract}
Leveraging temporal synchronization and association within sight and sound is an essential step towards robust localization of sounding objects. To this end, we propose a space-time memory network for sounding object localization in videos. It can simultaneously learn spatio-temporal attention over both uni-modal and cross-modal representations from audio and visual modalities. We show and analyze both quantitatively and qualitatively the effectiveness of incorporating spatio-temporal learning in localizing audio-visual objects. We demonstrate that our approach generalizes over various complex audio-visual scenes and outperforms recent state-of-the-art methods. Code and data can be found at \url{https://sites.google.com/view/bmvc2021stm}.
\end{abstract}

\section{Introduction}

Neurological evidence suggests that human understandings of scenes predominantly rely on the integration of visual and auditory cues \cite{fneur}. As humans, we form attention mechanisms to sounding sources by leveraging the temporal, cross-modal alignments between vision and sound, where understandings of the past tell us where and what to attend to next. For computational models, although there have been several developed sound source spatial localization frameworks \cite{Tian_2018_ECCV,Owens_2018_ECCV,qian2020multiple}, how much we gain from explicitly leveraging temporal correspondence that exists naturally in both videos and audios is yet to be explored.

However, considerations of temporal coherence are required to facilitate consistent understandings in complex scenes. Imagine a person playing a guitar in front of a wall of not-in-use guitars. In order to figure out which guitar is sounding and obtain stable localization results, we must take multiple timesteps into account. Hence, it is worthwhile to explore if learning temporal cues could benefit the localization of sounding objects in videos. 

To localize visual objects associated with specific sound sources, most of the previous works capture audio-visual spatial correspondence using similarities between audio and visual modalities~\cite{objs_that_sound,Owens_2018_ECCV,Hu_2019_CVPR}, cross-modal attention mechanisms~\cite{Senocak_2018_CVPR,Tian_2018_ECCV}, and sounding class activation mapping~\cite{qian2020multiple}. Nevertheless, these methods often identify sounding objects for static images, and audio-visual temporal coherence in videos is commonly ignored.
\begin{figure}[t]
\centering
\includegraphics[width=0.6\linewidth]{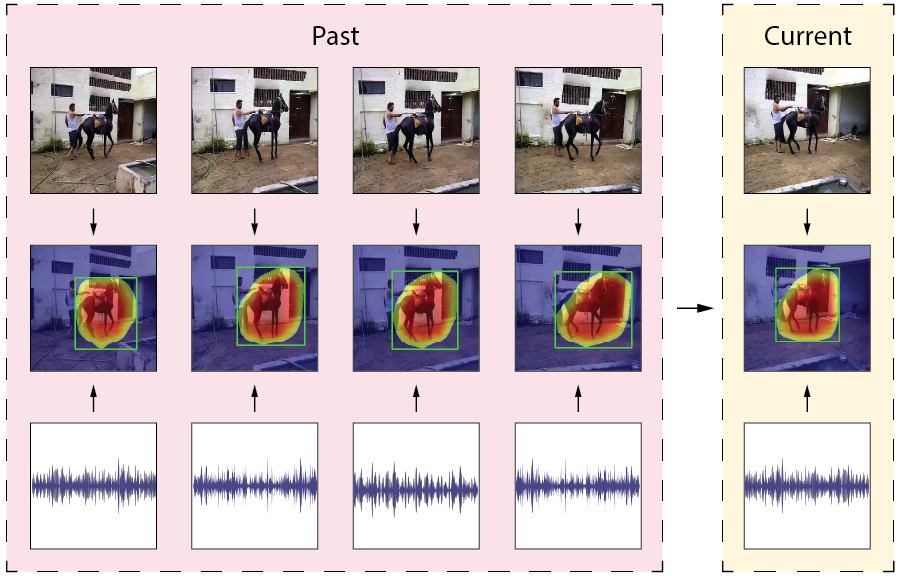}
\vspace{-3mm}
\caption{Our space-time memory network learns to attend to objects that currently sound by leveraging the temporal, cross-modal correspondence within sight and sound from the past. Here, given the frames of the two objects in motion and the sound of the horse walking, our model outputs stable and accurate localization results.}
\label{fig:teaser}

\end{figure}

Consequently, the essential question we must answer is how to design an efficient multi-modal deep neural network architecture that exploits the temporal coherence in visual frames and the corresponding audio segments. In this paper, we propose a spatio-temporal attention-based memory module, that can learn rich reference information from uni-modal as well as cross-modal (audio-visual) representations. With temporal memory updates, our approach is more robust against appearance and acoustic changes than the previous methods. It yields more temporally consistent localization results and can handle the absence of audio-visual events. In particular, to demonstrate the values of multi-modal temporal learning in sounding object localization, we resort to an easily affordable, weakly-supervised task in classifying the audio-visual event category of a given video segment.

Herein, our main contributions are: (1) we propose a novel space-time memory network that learns representations of sounding objects to promote robust localization performance, as illustrated in Figure \ref{fig:teaser}; (2) we validate the effectiveness of temporal learning in localizing sounding objects both quantitatively and qualitatively based on numerical benchmarks and visual interpretations; (3) we demonstrate that our approach generalizes over various complex audio-visual events and outperforms recent state-of-the-art methods.

\section{Related Work}

\subsection{Sounding Object Localization}
Sounding object localization refers to the task of localizing visual objects/scenes associated with specific sounds in videos. Early works resorted to mutual information \cite{nips1999} and canonical correlation analysis \cite{pixels_that_sound} to perform localization and segmentation on sounding pixels. Recent efforts have learned deep audio-visual models to localize sounding pixels, using audio-visual embedding similarities \cite{objs_that_sound,Owens_2018_ECCV,Hu_2019_CVPR}, cross-modal attention mechanisms \cite{Senocak_2018_CVPR,Tian_2018_ECCV}, vision-to-sound knowledge transfer \cite{Gan_2019_ICCV}, sounding class activation mapping \cite{qian2020multiple,hu2020discriminative}, and sounding object visual grounding~\cite{Tian_2021_CVPR}. While these methods work well on a single sound source in the simple audio-visual scenes, they lack temporal knowledge and predict audible regions solely based on the association of the current video frame with the corresponding audio segment. Most recently, Afouras et al.~\cite{Afouras20b} compute audio-visual cross-modal attention to spatially localize sounding regions. Moreover, they incorporate temporal learning into the visual modality, where they propose to use optical flow that is separately learned to aggregate information over time and group sound sources into audio-visual objects. Their model also relies on speech-oriented tasks and scenes, assuming objects (speakers) of fixed size. By contrast, not only does our spatio-temporal attention mechanism consider both uni-modal and cross-modal representations, but it is also learned in an end-to-end manner. Hence, different from the previous methods, with learnable space-time memory modules, our model can effectively leverage multi-modal contexts for localizing sounding objects and thus is capable of handling diverse and complex audio-visual objects.

\subsection{Audio-Visual Video Understanding}
The community has attracted an increasing amount of interest in recent years since synchronized audio-visual scenes are widely available in videos. In addition to localizing sound sources, a wide range of tasks have been proposed, including audio-visual sound separation~\cite{gao2018objectSounds, Zhao_2018_ECCV, Zhao_2019_ICCV, Gan_2020_CVPR, Tian_2021_robustness}, audio-visual action recognition~\cite{Gao_2020_CVPR,Kazakos_2019_ICCV,Korbar_2019_ICCV,Tian_2021_CVPR}, audio-visual event localization~\cite{Tian_2018_ECCV,dual_attn_matching}, audio-visual video captioning~\cite{Rahman_2019_ICCV,Tian_2019_CVPR_Workshops,watch_listen_describe}, embodied audio-visual navigation~\cite{Gan2020LookLA,chen20soundspaces}, audio-visual sound recognition~\cite{ijcai2020-78}, and audio-visual video parsing~\cite{tian2020avvp}. Our framework demonstrates that temporal learning facilitates better audio-visual understanding, which explicitly and subsequently benefits the localization performance.

\section{Proposed Method}

\begin{figure}[t]
    \centering
    \includegraphics[width=0.95\linewidth]{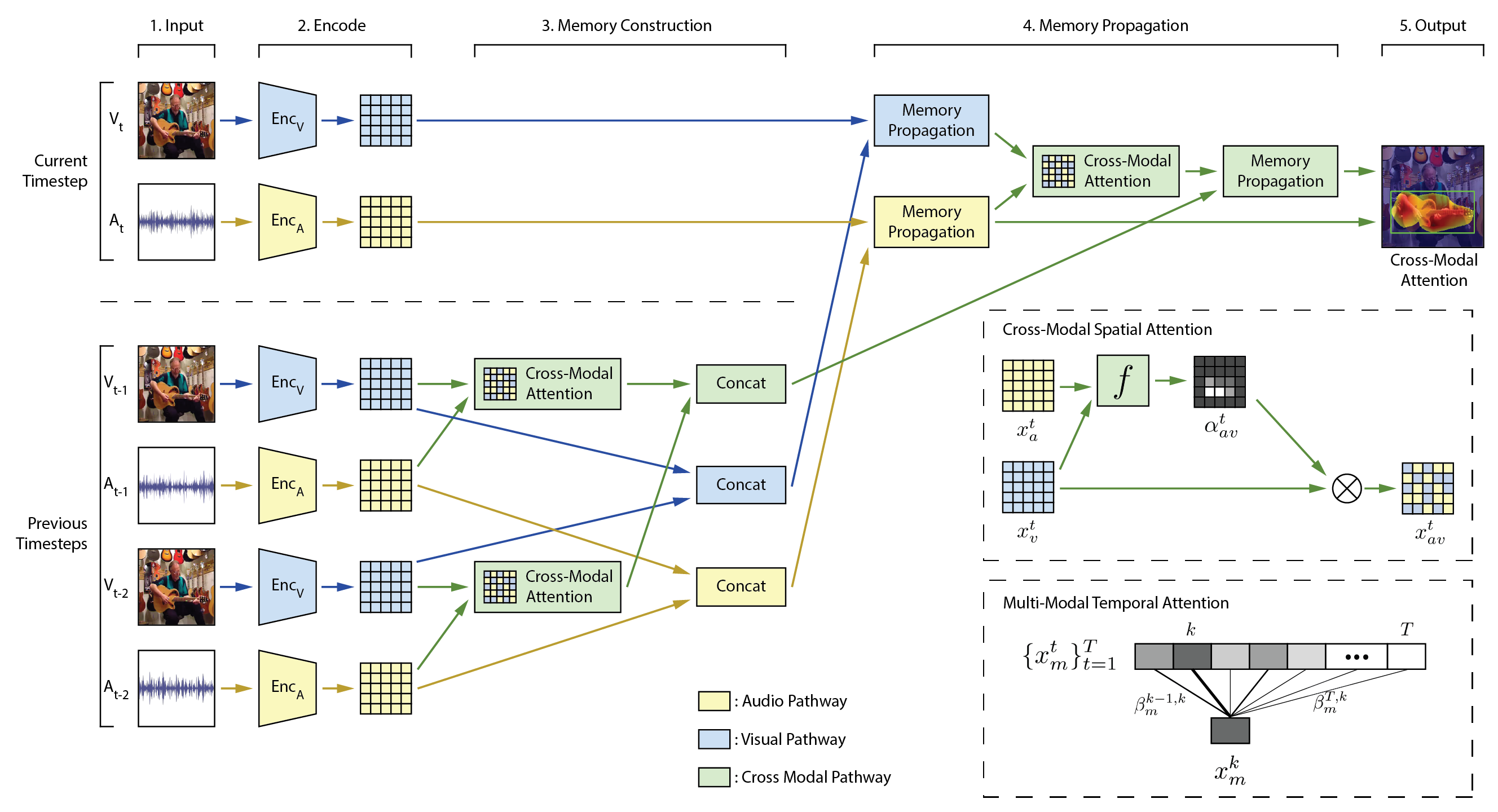}
    \caption{\textbf{Overview of the Space-Time Memory Network}: (1-2) extracts audio and visual features from inputs. (3) constructs one cross-modal memory module and two uni-modal memory modules. (4) first propagates the uni-modal memory, then computes cross-modal attention, and propagates the cross-modal memory. (5) computes spatial attention from the memory-propagated audio features to the memory-propagated audio-visual features. The outputs is an attention map that localizes sounding objects, which is used for downstream audio-visual event classification during training.}
    \label{fig:framework overview}

\end{figure}

Our framework (see Figure \ref{fig:framework overview}) consists of three modules: audio and visual feature extraction, memory construction and propagation, and sounding object localization. Given a video, it learns to attend to objects that sound in video frames. The spatio-temporal attention mechanisms are designed to leverage both uni-modal and cross-modal representations. It is supported by the idea of accumulating the past evidence into memory, which is then aggregated and propagated onto the current timestep. Through the task of audio-visual event classification, our model facilitates audio-visual understanding by learning spatio-temporal attention mechanisms that locate sounding objects.

\subsection{Audio-Visual Feature Extraction}
\def\vid{X_v^t}
\def\aud{X_a^t}
\def\viddim{\mathbb{R}^{H \times W \times 3}}
\def\auddim{\mathbb{R}^{M \times N \times 1}}
\def\vfdim{\mathbb{R}^{h \times w \times C}}
\def\afdim{\mathbb{R}^{m \times n \times C}}
\def\vfeat{x_v^t}
\def\afeat{x_a^t}

Consider $T$ audio-visual pairs $\{\vid, \aud\}^{T}_{t=1}$ as inputs, where $\vid \in \viddim, \aud \in \auddim$ denote the frame and its corresponding audio log-mel spectrogram at timestep $t$ respectively. Let $H, W, M, N$ denote the height, width, frequency, and time. For a given pair, we employ two convolutional encoders to project uni-modal features into C-dimensional joint audio-visual subspace for spatial-temporal attention in memory. Let $h, w$ and $ m, n$ denote the corresponding spatial dimensions of the feature maps. We obtain visual feature map $\vfeat \in \vfdim$ and audio feature map $\afeat \in \afdim$. This is reflected by steps 1 to 2 in Figure \ref{fig:framework overview}.

\subsection{Audio-Visual Memory Accumulation and Propagation}

To efficiently accumulate and aggregate the temporal audio-visual evidence, we propose to build memory modules for audio and visual representations both separately and jointly, which we collectively refer to as the \textit{Spatio-Temporal Memory Layer}. Here, we introduce the attention mechanisms in space and time, then describe how they form the memory layer.

\def\spatialattn{\alpha_{av}^t}
\def\avfeat{x_{av}^t}
\def\SpatialAttnDim{\mathbb{R}^{h \times w }}

\noindent \textbf{Cross-Modal Spatial Attention $f$.} 
We measure the similarity between the extracted uni-modal features $\vfeat$ and $\afeat$ at the given timestep $t$. We denote the cross-modal spatial attention function as $f(\afeat, \vfeat) = \spatialattn$, such that $f: \afdim \times \vfdim \mapsto \SpatialAttnDim$. For each position $(i, j)$ in $x_v^{t}$, the attention weight can be computed by: 
\begin{equation}
\begin{split}
    \label{eqn:spatial_att}
    \spatialattn(i, j) &= \frac{exp({\vfeat(i, j)}{\phi(\afeat)}^{\intercal})}{\sum_{i, j}exp({\vfeat(i, j)}{\phi(\afeat)}^{\intercal})}. \\
\end{split}
\end{equation}
Here, we adopt dot product as a generic choice to compute the spatial similarity and $\phi(\cdot)$ is a global pooling operation over the spatial dimension of its input.
We compute the cross-modal audio-visual features $\avfeat \in \vfdim$ by multiplying the learned spatial attention map with every channel $c \in \{1, ..., C\}$ in the visual feature map. Specifically, $\avfeat(i, j, c) = \spatialattn(i, j) * \vfeat(i, j, c)$.
For every pixel in a given frame $\vid$, we have thus found its affinity with the aligned audio input $\aud$ via cross-modal spatial attention $f$ on the feature level.

\def\temporalattn{\beta_{m}^{k,t}}
\def\memlength{\{1, ..., K\}}
\def\currfeat{x_m^t}
\def\prevfeat{x_m^k}
\def\memqkdim{\mathbb{R}^{h_m \times w_m \times C}}
\noindent \textbf{Multi-Modal Temporal Attention $g$.} Our memory formulation is generic and can be adapted to both uni-modal and cross-modal representations. Consider the modality $m \in \{a, v, av\}$ and the previous timestep $k \in \memlength$. We seek to measure the importance of $\prevfeat$ with respect to $\currfeat$ from the current timestep. We denote the multi-modal temporal attention function as $g(\currfeat, \prevfeat) = \temporalattn$, where $h_m, w_m$ denote the spatial dimensions of the feature maps of modality $m$, such that $g: \memqkdim \times \memqkdim \mapsto \mathbb{R}$.

We adopt the generic multi-head scaled dot-product attention from Vaswani et al. \cite{attention_is_all_u_need}, where we view $\currfeat$ as queries and $\prevfeat$ as keys and values. For each of the $L$ attention heads, we globally pool the input feature maps into feature vectors, and compute the set of attention weights over $K$ timesteps. Concretely, the multi-modal temporal attention function is formulated as:
\begin{equation}
\label{eqn:temporal_att}
    \beta^{k, t}_m = \frac{1}{L}\sum_{l=1}^{L}A(\phi(x^t_m)W^{query}_{l}, \phi(x^k_m)W^{key}_{l}) \\
\end{equation}
where $ A(I_{query}, I_{key}) = softmax(\frac{I_{query} I_{key}^T}{\sqrt{C}})$ denotes the attention function, $C$ corresponds to the number of dimensions for keys and queries, and $W_l^{query}, W_l^{key}$ are the learnable projections.

Having computed the set of attention weights, we propagate the memory to the current timestep $t$ to obtain $    \dot{x}_{m}^{t} = \beta_{m}^{t,t} * \currfeat + \sum_{k=1}^{K}{\temporalattn * \prevfeat}$.
Hence, the multi-modal temporal attention takes into account the importance of the past $k$ to the present $t$.

\def\memafeat{\dot{x}_a^t}
\def\memvfeat{\dot{x}_v^t}
\def\memavfeat{\dot{x}_{av}^t}
\def\twicememavfeat{\ddot{x}_{av}^t}

\noindent \textbf{Spatio-Temporal Memory Layer.} We now describe the algorithm that combines the two proposed attention mechanisms in space and time. Our memory layer first leverages the uni-modal association in time by applying the temporal attention $g$ to $\afeat$, $\vfeat$, respectively. It then considers the cross-modal association in space by applying the spatial attention $f$ to $\memafeat, \memvfeat$. Finally, the memory layer derives the cross-modal association in time by applying the temporal attention $g$ to $\memavfeat$. This is reflected by steps 3 to 4 in Figure \ref{fig:framework overview} and Algorithm \ref{algo: memory_layer}.

\begin{algorithm}[H]
    \caption{Spatio-Temporal Memory Layer Pseudocode}
    \label{algo: memory_layer}
    \begin{algorithmic}[1]
    \Procedure{MemoryLayerForward}{$x_a^t, x_v^t, \{(x_a^k, x_v^k, x_{av}^k)\}_{k=1}^K$} 
        \State $\memafeat = TemporalAttention(x_a^t, \{x_a^k\}_{k=1}^K)$ \Comment{Uni-Modal Temporal Attention}
        \State $\memvfeat = TemporalAttention(x_v^t, \{x_v^k\}_{k=1}^K)$ \Comment{Uni-Modal Temporal Attention}
        \State $\memavfeat = SpatialAttention(\memafeat, \memvfeat)$ \Comment{Cross-Modal Spatial Attention}
        \State $\twicememavfeat = TemporalAttention(\memavfeat, \{x_{av}^{k}\}_{k=1}^K)$ \Comment{Cross-Modal Temporal Attention}
        \State \textbf{return} $\twicememavfeat$, $\memafeat$
    \EndProcedure
    \end{algorithmic}
\end{algorithm}

\def\twicememavspatialattn{\ddot{\alpha}_{av}^t}
\subsection{Localizing sounding objects}

To extract audio-visual objects from various audio-visual events, we cannot impose an assumption on the sizes of the objects. We propose two post-processing approaches, using contour detection and pre-trained object-proposal-networks, respectively.

Given outputs $\memafeat, \twicememavfeat$ from the memory layer, we compute the cross-modal spatial attention map $\twicememavspatialattn=f(\memafeat, \twicememavfeat)$, which we view as the final sounding object localization map. This is reflected by step 5 in Figure \ref{fig:framework overview}. We generate bounding boxes by applying Otsu's threshold~\cite{otsu_threshold} and contour detection to the normalized spatial attention map. Alternatively, the attention map $\twicememavspatialattn$ can also be incorporated into robust object instances generated by out-of-the-box object proposal methods. Given frame $\vid$, we extract $N$ object instances using a region-proposal-network (RPN). We convert them into binary masks $\{m_n^t\}_{n=1}^N$, with $1$ indicating the instance and $0$ otherwise. We calculate the individual score of each box $S_n^t$ as the weighted sum between $m_n^t$ and $\twicememavspatialattn$, or $S_n^t = \sum_{i, j}{m_n^t(i, j) * \twicememavspatialattn(i, j)}$, and apply non-maximum suppression (NMS) to filter overlapping boxes.

\subsection{Learning Spatio-Temporal Attention Mechanisms}

We utilize the easily affordable, weakly-supervised classification task on audio-visual event categories to learn the proposed spatio-temporal attention mechanisms. We fuse the outputs from the memory layer, including the uni-modal audio feature vector $\memafeat$ and the cross-modal audio-visual feature map $\twicememavfeat$, to obtain a joint representation. In particular, we sum $\twicememavfeat$ over its spatial dimensions, since it is already weighted by the spatial attention $f$, and concatenate the result with $\memafeat$. This gives us the final output of our network at timestep $t$, denoted as $\mathcal{O}^t \in \mathbb{R}^{2C}$. Concretely, $\mathcal{O}^t = [\sum_{i, j}\twicememavfeat(i, j); \memafeat]$, where $[...; ...]$ denotes concatenation. This joint audio-visual representation is used to estimate the audio-visual event category for the given video segment using a multilayer-perceptron (MLP) and the cross-entropy loss.

\section{Experiments}
\subsection{Datasets}
To carefully and temporally evaluate the localization performance, we emphasize three aspects when tailoring dataset: the scope of the sounding object categories covered, the scale of the testing examples contained, and the number of frames per video densely annotated. Therefore, we use AVE dataset~\cite{Tian_2018_ECCV} for scope and AudioSet-Instrument dataset~\cite{AudioSet} for scale. For both datasets, we densely annotated frames in the test videos.

\noindent \textbf{AVE: Audio-Visual Event Dataset.}
AVE dataset~\cite{Tian_2018_ECCV} consists of 4143 10-second video clips that are labeled with the audio-visual event, covering 28 categories. Although the dataset is also temporally labeled with audio-visual event boundaries, to demonstrate the effectiveness of our weakly-supervised learning framework, we include the whole length of every video. We adopt the original data split, including 3339 videos for training, 402 for validation, and 402 for testing.

\noindent \textbf{AudioSet-Instrument.} AudioSet-Instrument dataset is a subset of AudioSet~\cite{AudioSet} that consists of 101076 10-second video clips, spanning across 15 instrument categories. It contains challenging video clips, where many are of poor quality with multiple sound sources. We use 100221/424/424  for training/validation/testing.

\noindent \textbf{Annotations.} We annotated test sets in AVE and AudioSet-Instrument, where we created bounding boxes for sounding objects in the duration of the audio-visual event. For reproducibility, we will release the annotations with source code.

\subsection{Implementation Details}
\noindent \textbf{Data Preprocessing.} During training, we randomly sample one second and extract its corresponding audio-visual pair $\{(\vid, \aud)\}^{T}_{t=1}$, where $T$ denotes the number of frames, as well as the number of audio segments, we sample per second. In practice, we use $T=4$. During the evaluation, $T$ corresponds to the number of frames that are annotated in the given video. Final input frames are of spatial dimensions $256 \times 256$. During training, this is achieved by resizing the image by a scale of $1.1$ and randomly crop it to the desired size. During the evaluation, every frame is directly resized to the target dimensions. Audio inputs are re-sampled to 16 kHz mono and its corresponding spectrogram is computed through Short-Time Fourier Transform with a Hann window size of 25ms and a hop length of 10ms. For every input frame, we use the output spectrogram of its temporally-aligned audio segment.

\noindent \textbf{Encoder.} We use ResNet-152 \cite{ResNet} to extract 2048-D per-frame visual features and VGGish \cite{VGGish} to extract 512-D per-segment audio features, they are then each projected to a 256-D joint feature space by a stack of 1x1 convolutions with ReLU non-linearity.

\noindent \textbf{Memory Module.} We use Multi-head scaled dot product attention with 256-D embedding dimension and 1 head.

\noindent \textbf{Sounding Object Extraction.} To localize objects that sound from the cross-modal spatial attention map, we use bilinear interpolation to resize to $256\times256$. We then normalize the attention map by subtracting the minimum and dividing by the maximum, resulting in a lower bound of $0$ and an upper bound of $1$. We use Otsu's threshold and contour detection to extract bounding boxes for experimental comparison and ablation study. For the RPN-based extraction method demonstrated in Figure \ref{fig: grounding}, we used a Faster R-CNN model with a ResNet-50-FPN backbone \cite{faster_rcnn,ResNet}, pretrained on COCO train2017.


\noindent \textbf{Hyperparameters.} In our experiments, we use batch size 128 and 30 epochs. Our framework is trained with Adam optimizer with the initial learning rate of $10^{-4}$ on four NVIDIA 1080Ti GPU.

\begin{figure}[t]
    \centering
    \includegraphics[width=\linewidth, height=0.35\textheight]{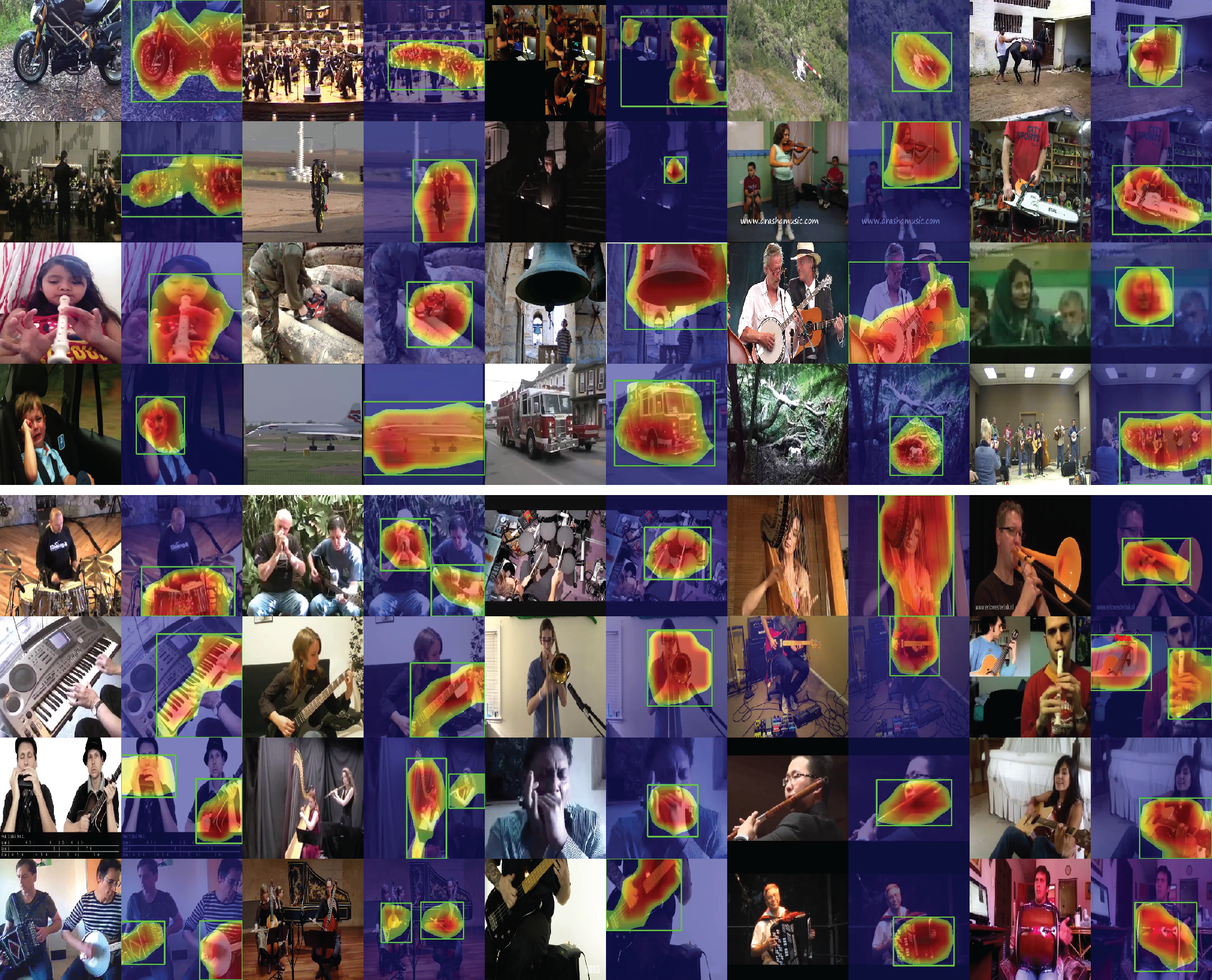}
    \vspace{-6mm}
    \caption{\textbf{Qualitative Visualizations}: We show localization results from the two datasets. Top four rows contain results from the AVE dataset. Bottom four rows contain results from the Audioset-Instrument dataset. Bounding boxes are extracted from the cross-modal spatial attention maps, using Otsu's threshold and contour detection.}
    \label{fig:results_grid}

\end{figure}

\noindent \textbf{Evaluation Metrics.}
To evaluate the localization performance of our framework, we employ Intersection over Union (IoU) for the box-level localization performance and Consensus Intersection over Union (cIoU) proposed in \cite{Senocak_2018_CVPR} for the pixel-level localization performance. We now expand on details of cIoU calculations. Given an annotated frame, we convert the ground truth bounding boxes $\{b_j\}_{j=1}^{N}$ into a binary ground truth map $\mathbf{g}$, where $1$ indicates that a pixel is sounding and $0$ otherwise. Given predicted location map $\alpha$, we define cIoU under threshold $\tau$ as: $ cIoU(\tau) = \frac{\sum_{i\in \mathcal{A}(\tau)} \mathbf{g}_i}{\sum_{i}{\mathbf{g}_i} + \sum_{i \in \mathcal{A}(\tau) - \mathcal{G}}{1}}$.
Here, $i$ is the pixel index of the map. $\mathcal{A}(\tau) = \{i | \alpha_i > \tau\}$ denotes the set of pixels with attention intensity higher than the threshold, and $\mathcal{G} = \{i | \mathbf{g}_i > 0 \}$ represents the set of pixels annotated as positive. We use 0.5 as the cIoU threshold in our experiments.

\begin{figure}[t]
    \centering

    \subfigure[Input Frames]{
        \includegraphics[width=2.4cm, height=1.6cm]{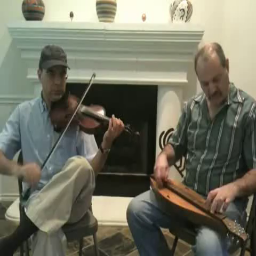}  
        \includegraphics[width=2.4cm, height=1.6cm]{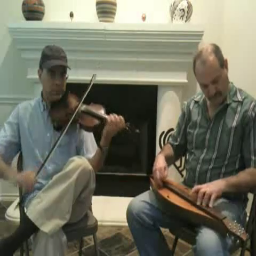}  
        \includegraphics[width=2.4cm, height=1.6cm]{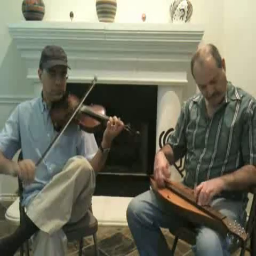} 
        \includegraphics[width=2.4cm, height=1.6cm]{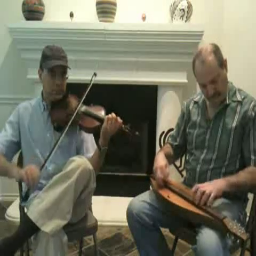}  
        \includegraphics[width=2.4cm, height=1.6cm]{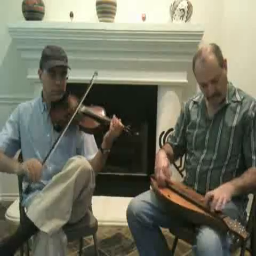}
    }\vspace{-3mm}

    \subfigure[Tian et al.~\cite{Tian_2018_ECCV}]{
        \includegraphics[width=2.4cm, height=1.6cm]{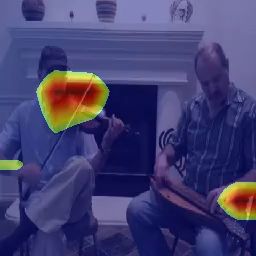}  
        \includegraphics[width=2.4cm, height=1.6cm]{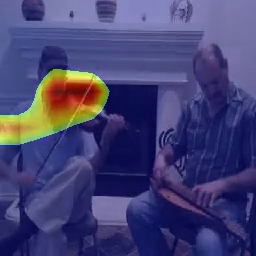} 
        \includegraphics[width=2.4cm, height=1.6cm]{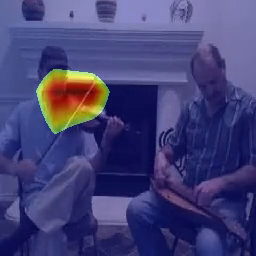}  
        \includegraphics[width=2.4cm, height=1.6cm]{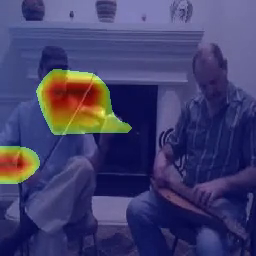}  
        \includegraphics[width=2.4cm, height=1.6cm]{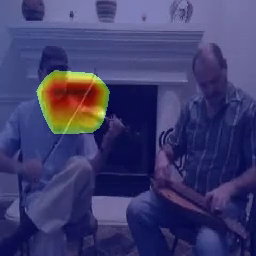}
    }\vspace{-3mm}
    
    \subfigure[Qian et al.~\cite{qian2020multiple}]{
        \includegraphics[width=2.4cm, height=1.6cm]{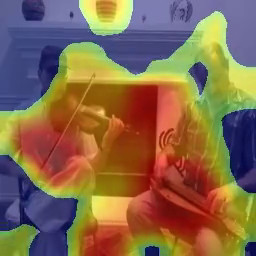}  
        \includegraphics[width=2.4cm, height=1.6cm]{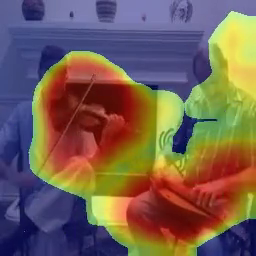} 
        \includegraphics[width=2.4cm, height=1.6cm]{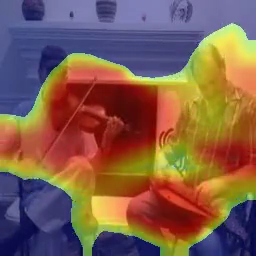} 
        \includegraphics[width=2.4cm, height=1.6cm]{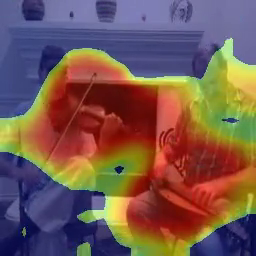}  
        \includegraphics[width=2.4cm, height=1.6cm]{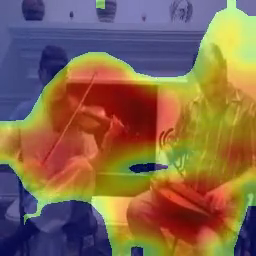}
    }\vspace{-3mm}
    
    \subfigure[Hu et al.~\cite{hu2020discriminative}]{
        \includegraphics[width=2.4cm, height=1.6cm]{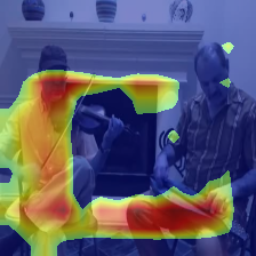}  
        \includegraphics[width=2.4cm, height=1.6cm]{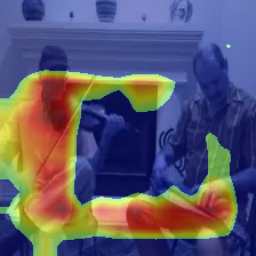} 
        \includegraphics[width=2.4cm, height=1.6cm]{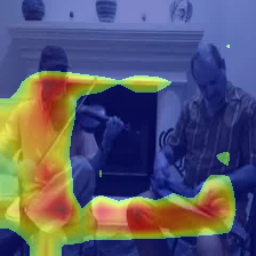} 
        \includegraphics[width=2.4cm, height=1.6cm]{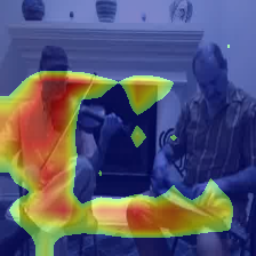}  
        \includegraphics[width=2.4cm, height=1.6cm]{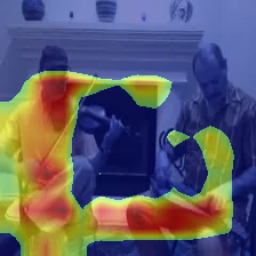}
    }\vspace{-3mm}
    
    \subfigure[Our model]{
        \includegraphics[width=2.4cm, height=1.6cm]{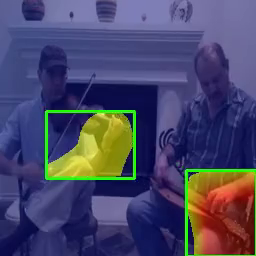}
        \includegraphics[width=2.4cm, height=1.6cm]{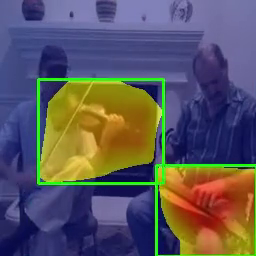}
        \includegraphics[width=2.4cm, height=1.6cm]{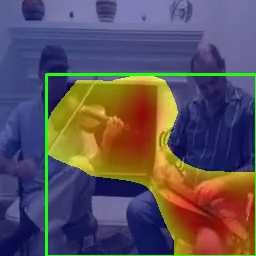}
        \includegraphics[width=2.4cm, height=1.6cm]{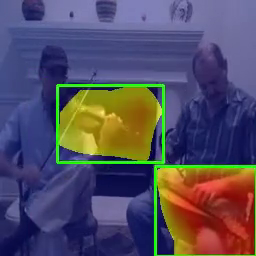}  
        \includegraphics[width=2.4cm, height=1.6cm]{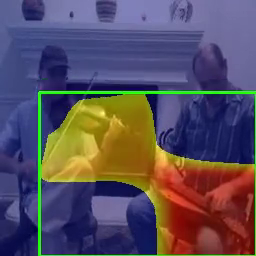}
    }
    
    \caption{\textbf{Experimental Comparison}: We qualitatively compare the localization performance of our framework with two recent methods by Tian et al.~\cite{Tian_2018_ECCV} and Qian et al.~\cite{qian2020multiple}.}
    \label{fig: experimental comparison}

\end{figure}

\begin{table}[h]
    \centering
    \resizebox{0.7\textwidth}{!}{%
    \begin{tabular}{c|cccc}
        \toprule
        Method & \multicolumn{2}{c}{AVE} & \multicolumn{2}{c}{Audioset-Instrument} \\
               &    cIoU@0.5  & IoU@0.5    & cIoU@0.5  & IoU@0.5   \\
        \midrule
        Tian et al.~\cite{Tian_2018_ECCV}             & 11.73       & 18.81     & 25.11               & 33.20              \\
        Owens et al.~\cite{Owens_2018_ECCV}          & 13.96       & 22.64     & 21.79               & 38.08    \\
        Qian et al.~\cite{qian2020multiple}          & 24.16       & 16.82     & 20.16               & 31.31             \\
        Hu et al.~\cite{hu2020discriminative}        & {21.25}       & {33.05}     & {33.74}               & {40.02}             \\
        Ours                    & \textbf{37.78}      & \textbf{37.50}      & \textbf{51.06}               & \textbf{56.49}             \\
        \bottomrule
    \end{tabular}
    }
    \caption{Localization results of different methods on AVE and Audioset-Instrument datasets. All methods are evaluated by IoU@0.5 and cIoU@0.5. The top-1 result in each column is highlighted.}
    \label{tbl: experimental comparison}
\end{table}


\subsection{Results}

In Figure \ref{fig:results_grid}, we illustrate sounding object localization results on AVE and Audioset-Instrument datasets. We observe that our model is capable of correctly discovering sounding regions for a wide range of sound sources in challenging unconstrained videos.

\noindent \textbf{Experimental Comparison.} To further validate the effectiveness of our space-time memory network, we compare it with three recent methods: Owens and Efros~\cite{Owens_2018_ECCV}, Tian et al.~\cite{Tian_2018_ECCV}, Qian et al.~\cite{qian2020multiple}, and Hu et al.~\cite{hu2020discriminative}\footnote{Afouras et al.~\cite{Afouras20b} was not compared since their framework is trained on speech-oriented downstream tasks and imposes an assumption on the size of the sounding object.}. We demonstrate the quantitative results in Table \ref{tbl: experimental comparison} and the qualitative results in Figure \ref{fig: experimental comparison}\footnote{More results can be found in our supplementary material.}. We find that our framework outperforms the compared approaches on AVE and Audioset-Instrument datasets both quantitatively and qualitatively, which substantiates the benefits of the proposed
space-time memory network in localizing dynamic audio-visual objects.

\begin{figure}[t]
    \centering
    \includegraphics[width=\textwidth]{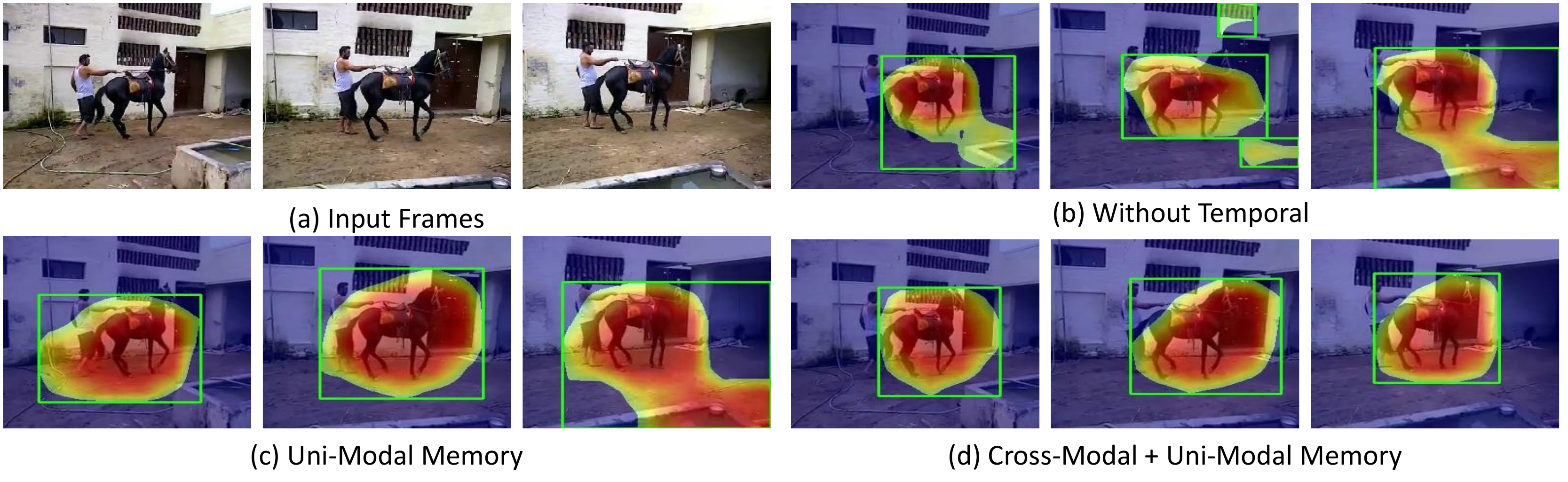}
    \vspace{-6mm}
    \caption{\textbf{Ablation Study}: Visualizations of sounding object localization from the three ablative groups. Here, only horse is making audible sounds.}
    \label{fig: ablation study}

\end{figure}

\noindent \textbf{Ablation Study.}
To evaluate the effectiveness of our proposed memory module, we conduct ablation study for the following models: (1) \textit{Cross-Modal Memory + Uni-Modal Memory} (C+U): we employ both uni-modal and cross-modal temporal learning modules. (2) \textit{Uni-Modal Memory} (U): we remove the cross-modal memory module, propagating memory on a uni-modal level only. (3) \textit{Without Temporal}: we remove both uni-modal and cross-modal memory modules. Without temporal learning, the baseline model associates frames and sounds on single timesteps. We show the quantitative results in Table \ref{tbl: ablation study} and the qualitative results in Figure \ref{fig: ablation study}. We find that (1) outperforms the other two ablative groups numerically and demonstrates significantly more robust visualization results.

\begin{table}[h]
    \centering
    \resizebox{0.7\textwidth}{!}{%
    \begin{tabular}{c|cccc}
        \toprule
        Method & \multicolumn{2}{c}{AVE} & \multicolumn{2}{c}{Audioset-Instrument} \\
              & cIoU@0.5    & IoU@0.5   & cIoU@0.5            & IoU@0.5           \\
        \midrule
        
        Without Temporal                & 34.81       & 33.82     & 44.30           & 51.57             \\
        U              & 36.41       & 35.23     & 48.96           & 53.19              \\
        C+U             & \textbf{37.78}       & \textbf{37.50}     & \textbf{51.06}           & \textbf{56.49}    \\
        \bottomrule
    \end{tabular}%
    }
    \caption{\textbf{Ablation Study}: Localization results of the ablative groups on AVE and Audioset-Instrument datasets. All methods are evaluated by IoU@0.5 and cIoU@0.5. The top-1 result in each column is highlighted.}
    \label{tbl: ablation study}
\end{table}



\begin{figure}[h]
    \centering
    \subfigure{
        \includegraphics[width=2.4cm, height=1.6cm]{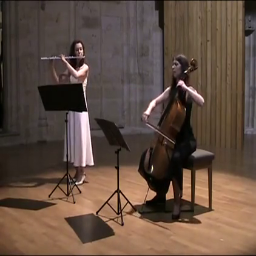} 
        \includegraphics[width=2.4cm, height=1.6cm]{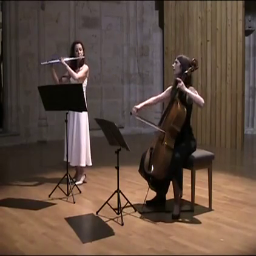} 
        \includegraphics[width=2.4cm, height=1.6cm]{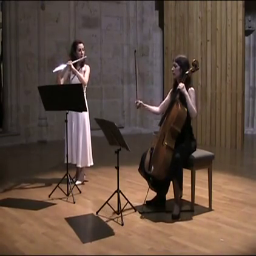}
        \includegraphics[width=2.4cm, height=1.6cm]{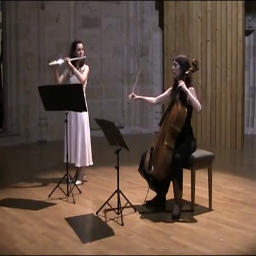} 
        \includegraphics[width=2.4cm, height=1.6cm]{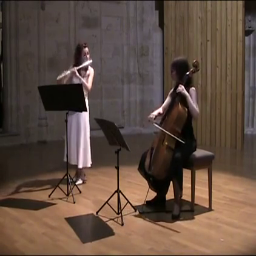} 
    }\vspace{-3mm}
    
    \subfigure{
        \includegraphics[width=2.4cm, height=1.6cm]{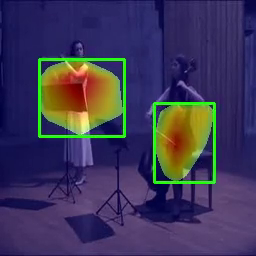} 
        \includegraphics[width=2.4cm, height=1.6cm]{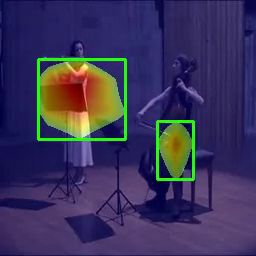} 
        \includegraphics[width=2.4cm, height=1.6cm]{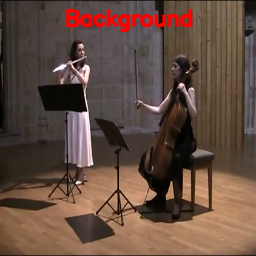} 
        \includegraphics[width=2.4cm, height=1.6cm]{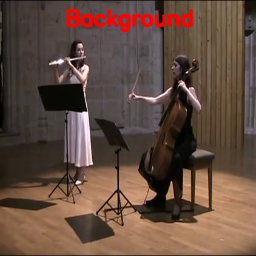} 
        \includegraphics[width=2.4cm, height=1.6cm]{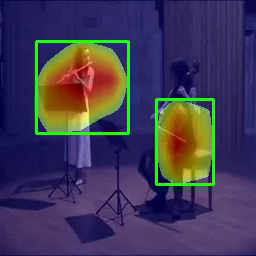} 
    }
    \caption{\textbf{Handling the absence of audio-visual events:} In this challenging example (top), both performers stop playing between the third and the fourth steps, and resume on the fifth step. We show our model performance (bottom), where background class is predicted following an absence of audio-visual events.}
    \label{fig: absence handling}
\end{figure}

\noindent \textbf{Handling the absence of audio-visual events.}
Given that not all audio-visual segments contain audio-visual events, we further demonstrate the robustness of our model in a challenging example in Figure \ref{fig: absence handling}. While the cello performer has lifted her bow up, it is impossible to tell whether the flute is sounding merely from sight. Our framework identifies the non-sounding frames by resorting to sound.

\begin{figure}[!h]
    \centering     
    \subfigure[Attention Output]{
        \includegraphics[width=0.3\linewidth, height=0.2\linewidth]{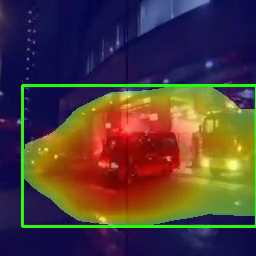}
    }
    \subfigure[Box Proposals]{
        \includegraphics[width=0.3\linewidth, height=0.2\linewidth]{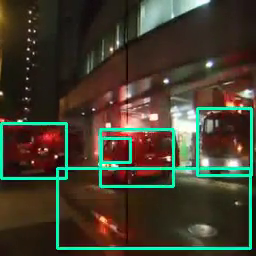}
    }
    \subfigure[AV Grounding]{
        \includegraphics[width=0.3\linewidth, height=0.2\linewidth]{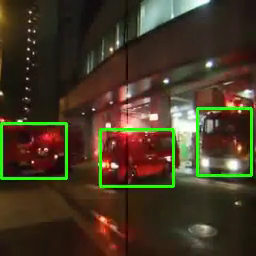}
    }
    \caption{\textbf{Audio-Visual Object Grounding:} Using a pretrained RPN, our grounding approach can further refine the localization performance of our framework. Here, we show (a) the final model output followed by contour detection, (b) the extracted box proposals, and (c) the audio-visual object grounding results.}
    \label{fig: grounding}
\end{figure}

\noindent \textbf{Audio-Visual Object Grounding.}
Following the RPN-based approach to extract sounding objects, we demonstrate how our cross-modal attention map can be incorporated to further refine localization performance in Figure \ref{fig: grounding}.
\vspace{-3mm}


\section{Conclusion}
In this paper, we investigate the effectiveness of multi-modal temporal learning in localizing audio-visual objects. We propose a novel space-time memory framework to address the problem. Results from experimental comparison and ablation study support our claim both objectively and subjectively that multi-modal temporal learning is crucial for robust sounding object localization performance.

\bigskip

\noindent \textbf{Acknowledgements}: We would like to thank the anonymous reviewers for the constructive comments. This work was supported in part by NSF 1741472 and 1909912. The article solely reflects the opinions and conclusions of its authors but not the funding agents.

\clearpage
\bibliography{bmvc_final}
\end{document}